\documentclass[conference]{IEEEtran}
\IEEEoverridecommandlockouts
\usepackage{cite}
\usepackage{amsmath,amssymb,amsfonts}
\usepackage{algorithmic}
\usepackage{graphicx}
\usepackage{textcomp}
\usepackage{xcolor}
\def\BibTeX{{\rm B\kern-.05em{\sc i\kern-.025em b}\kern-.08em
    T\kern-.1667em\lower.7ex\hbox{E}\kern-.125emX}}
\begin{document}
\IEEEoverridecommandlockouts
\IEEEpubid{\begin{minipage}[t]{\textwidth}\ \\[10pt]
        \centering\normalsize{978-1-6654-0945-2/21/\$31.00 \copyright 2021 IEEE}
\end{minipage}}

\title{Morphological Classification of Galaxies Using SpinalNet\\

}

\author{\IEEEauthorblockN{Dim Shaiakhmetov}
\IEEEauthorblockA{\textit{Department of Computer Science} \\
\textit{Ala-Too International University}\\
Bishkek, Kyrgyzstan \\
pearlice@mail.ru}

\hspace{8.0cm}
\and
\IEEEauthorblockN{Remudin Reshid Mekuria}
\IEEEauthorblockA{\textit{Department of Computer Science} \\
\textit{Ala-Too International University}\\
Bishkek, Kyrgyzstan \\
remudin@alatoo.edu.kg}
\\

\hspace{9.85cm}
\and
\IEEEauthorblockN{Ruslan Isaev}
\IEEEauthorblockA{\textit{Department of Computer Science} \\
\textit{Ala-Too International University}\\
Bishkek, Kyrgyzstan \\
ruslan.isaev@iaau.edu.kg}

\hspace{8.0cm}
\and

\IEEEauthorblockN{Fatma Unsal}
\IEEEauthorblockA{\textit{Department of Computer Science} \\
\textit{Ala-Too International University}\\
Bishkek, Kyrgyzstan \\
fatma.unsal@iaau.edu.kg}
}

\maketitle

\begin{abstract}
Deep neural networks (DNNs) with a step-by-step introduction of inputs, which is constructed by imitating the somatosensory system in human body, known as SpinalNet have been implemented in this work on a Galaxy Zoo dataset. The input segmentation in SpinalNet has enabled the intermediate layers to take some of the inputs as well as output of preceding layers thereby reducing the amount of the collected weights in the intermediate layers. As a result of these, the authors of SpinalNet reported to have achieved in most of the DNNs they tested, not only a remarkable cut in the error but also in the large reduction of the computational costs. Having applied it to the Galaxy Zoo dataset, we are able to classify the different classes and/or sub-classes of the galaxies. Thus, we have obtained higher classification accuracies of 98.2, 95 and 82 percents between elliptical and spirals, between these two and irregulars, and between 10 sub-classes of galaxies, respectively.
\end{abstract}


\section{Introduction}
Galaxies are gravitationally bound objects in the universe composed of stellar objects, gases and dust particles which are also filling up the space between them, as well as the dark matter which is a quite less understood type of matter from which they are largely made. The evolution of these objects, which is believed to have formed more than $\sim 10^{10}$ years ago, together with their visual appearance (shape, distribution of matter, and their structure) is expected to provide quite valuable information about their composition and their evolutionary changes. Categorizing galaxies into different classes is quite significant because astrophysicists routinely employ enormous databases of data to test existing ideas or generate new hypotheses to explain the physical processes that drive galaxies, star formation, and better understanding of the nature of the universe. Thus, galaxy morphology can be considered a basic quantity not only for obtaining all-encompassing information on the evolution of galaxies, but also for a wide range of science in observational cosmology (see for example \cite{Martin2020} and references therein).
 
A Galaxy Zoo initiative has emerged out of the need of astronomers from Oxford University in order to categorize galaxies according to their morphological classes, to better comprehend galactic dynamics \cite{GalaxyZooWebPage}. Galaxy Zoo adopted a unique approach in bringing astronomy to the general public, where they will log on and assist in the classification of a galaxies. There have been 4 different versions of the Galaxy Zoo projects. The first was concerned with determining if a galaxy was elliptical, spiral (together with its orientation), or the result of a merger of two galaxies. Galaxy Zoo 2 for instance has requested more information on bright or most prominent SDSS galaxies \cite{Willett2013}. These thorough categories include (among other things) bulge size measurements, the presence of bars, and the magnitude of the bulge. Kaggle challenge has used a data obtained from this segment.


Various attempts have been made to examine galaxies and categorize them into various shapes.
The authors of \cite{Calleja2004} for example have applied a set of uniform ensembles of classifiers which makes use of neural networks and a locally weighed regression technique. The later have been used for an easy extraction of features from the image datasets. They have reported that having pre-processed the galaxy images, they went on employing the principal component analysis which is not only effective in minimizing the dimensionality of the data but also in distilling quit useful information from them. The homogeneous ensemble of locally weighted regression delivers the greatest results, with over 95 percent accuracy when considering only the two classes of galaxies, namely elliptical and spirals, and over 91 percent when irregulars are also considered.

The author of \cite{Kalvankar2020} have implemented a deep neural networks (DNN) architecture which has a fixed set of scaling coefficients known as EfficientNets for morphological classifications of galaxies. They have used  $\sim 8.0 \times 10^{4}$ galaxy images which are obtained from Galaxy Zoo 2 datasets which were made available for Kaggle competition. They were able to classify galaxies into a total of 7 morphological classes. These are, 3 categories of elliptical shapes (i.e., completely, in-between and cigar shaped) smooths, 2 categories of spiral shapes (i.e., barred and unbarred) spirals and single categories of lenticulars and irregulars. An accuracy of $~94\%$ and an F1 score of 0.8857 is reported to have been achieved with the implementation of EfficientNets.

In their attempt to develop an automated morphological classification of galaxies, authors of \cite{Cavanagh2021} have trained and validated a number of convolutional neural network (CNN) architectures. They have applied them to more than 10 thousands of images of visually-classified Sloan Digital Sky Survey (SDSS) objects to classify them into both 3- morphological classes (i.e., elliptical, lenticular, spiral) and 4- morphological classes (these three and irregulars/miscellaneous). They claim to have developed a novel CNN architecture that outperforms previous models in both 3- and 4-way classification, with overall classification accuracies of 83\% and 81\%, respectively. They have compared the accuracies of binary classifications across all of the above mentioned four classes, finding that elliptical and spirals are the easiest to discern ($>$98 percent accuracy), while spirals and irregulars are the most difficult to distinguish (78 percent accuracy). They have investigated to understand the plausible physical reasons for those images which are classified incorrectly, for example most of the lenticular galaxies which were incorrectly classified to ellipticals were having higher stellar masses, similar to other trends which has also been mentioned. In additional to these, they have implemented the same CNN to a small sample of Galaxy Zoo datasets in order to classify them into the above-mentioned morphological classes. And they were able to obtain an accuracy of 92\% (for binary), 82\% (for 3-way) and 77\% (for 4-way) classifications.

Research focusing on similar aspect has also been using rather machine learning instead of deep learning approaches we have mentioned until now. For instance, the author of reference \cite{Manda2010} have applied an ensemble of machine learning approaches for various objects in the SDSS data release 6 which were classified by the Galaxy Zoo project in order to classify the galaxies into three morphological classes, namely early types, spirals and point sources/artefacts. The authors have also concluded that using machine-learning algorithms to perform morphological classification for the next generation of wide-field imaging surveys is quite promising, and that the Galaxy Zoo catalogue would continue serving as an essential training set for this purpose. Following that various works have picked these datasets and conducted a galaxy classification research's some of which include \cite{Kuminski2014, Variawa2020, Mike2021}.

\section{Morphological  Galaxy  Classification}

\begin{figure}[b]
\centering
\includegraphics[width=0.43\textwidth]{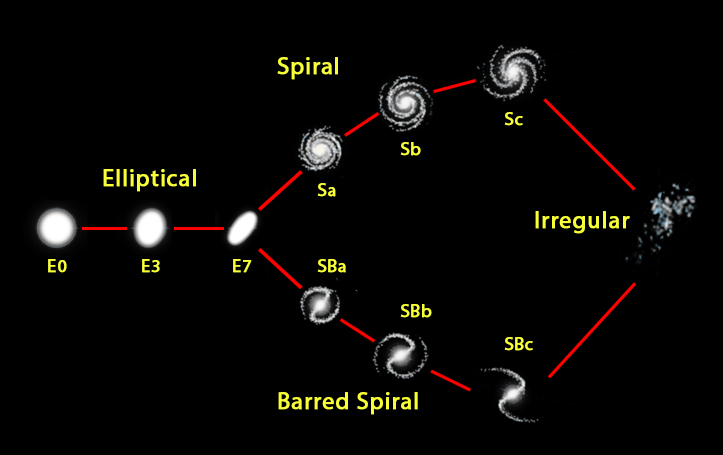} 

\caption{Hubble Tuning Fork.}
\label{fig1}
\end{figure}

In general, a system that is widely used by astronomers in order to classify galaxies into various classes based on their structure and appearance is what is commonly referred to as morphological galaxy classification. The most common classification scheme is the system devised by Sir Edwin Hubble in 1936 which is shown in Fig. \ref{fig1}. Hubble's original classifications include the following (i.e., also they were modified and extended later on by others to include more types):

(i) elliptical galaxies: (E0, E1, E2, E3, E4, E5, E6, E7)

(ii) spirals (Sa, Sb, Sc), barred spiral (SBa, SBb, SBc) and

(iii) irregulars.

This scheme is commonly referred to as the "Hubble Tuning Fork" \cite{Hubble}.


In this work we will classify galaxies by proceeded in the following manner:  first, we used 2-classes classifications between that of elliptical and spirals. Followed by the 3-classes by having introduced the irregulars. Lastly, we used 10-classes classifications namely, E0, E3, E7, Sa, Sb, Sc, SBa, SBb, SBc, Irr. Therefore, the contribution of this paper is to outline that having used SpinalNet which uses a DNN with a gradual input is an efficient technique for galaxy morphological (image) classifications as the higher and improved values of accuracies ($\sim$98.2\%) would exhibit. 

The remaining parts of the paper are organized as follows: in the next part (Section III), the experimental data used along with its pre-processing is briefly presented. In Section IV the implementation of the the SpinalNet is explained on the Galaxy Zoo datasets used in this work. In Section V a complete experimental results in respect to different classification accuracy measurements and confusion matrix for the test data are presented along with the corresponding discussions. A conclusion of this research work will be presented in Section VI.


\section{Materials and Methods}

In this research project, we have first classified galaxy image dataset into the three main morphological classes namely, elliptical, spirals and irregulars using SpinalNet, a deep neural network (DNN) code with a gradual input. The dataset was obtained from Kaggle \cite{GalaxyZooWebPage}. Imageset taken from dataset was categorized into three (3) respective classes with images divided into part for training and part for the validations. First, we split the dataset in a 70/30 proportion for training and testing sets, respectively. The training folder has been further divided into subfolders for the corresponding morphological classes. Similarly, the testing folder is also further divided into subfolders for the corresponding morphological classes. Following similar procedure, we have also added our experiment in classifying the galaxy image datasets into two (2) and ten (10) morphological classes. 

In order to perform an image classification, we put the image datasets distributed to various classes corresponding to their folders. We take the raw data from Galaxy Zoo project datasets and perform a data preparation by removing images which do not pass manually configured threshold. We ought to say that even after having image dataset cleared out, we need to manually remove some images that are not correctly fitting the requirement of the individual galaxy classes. 

\begin{figure}
\centering
\includegraphics[width=0.47\textwidth]{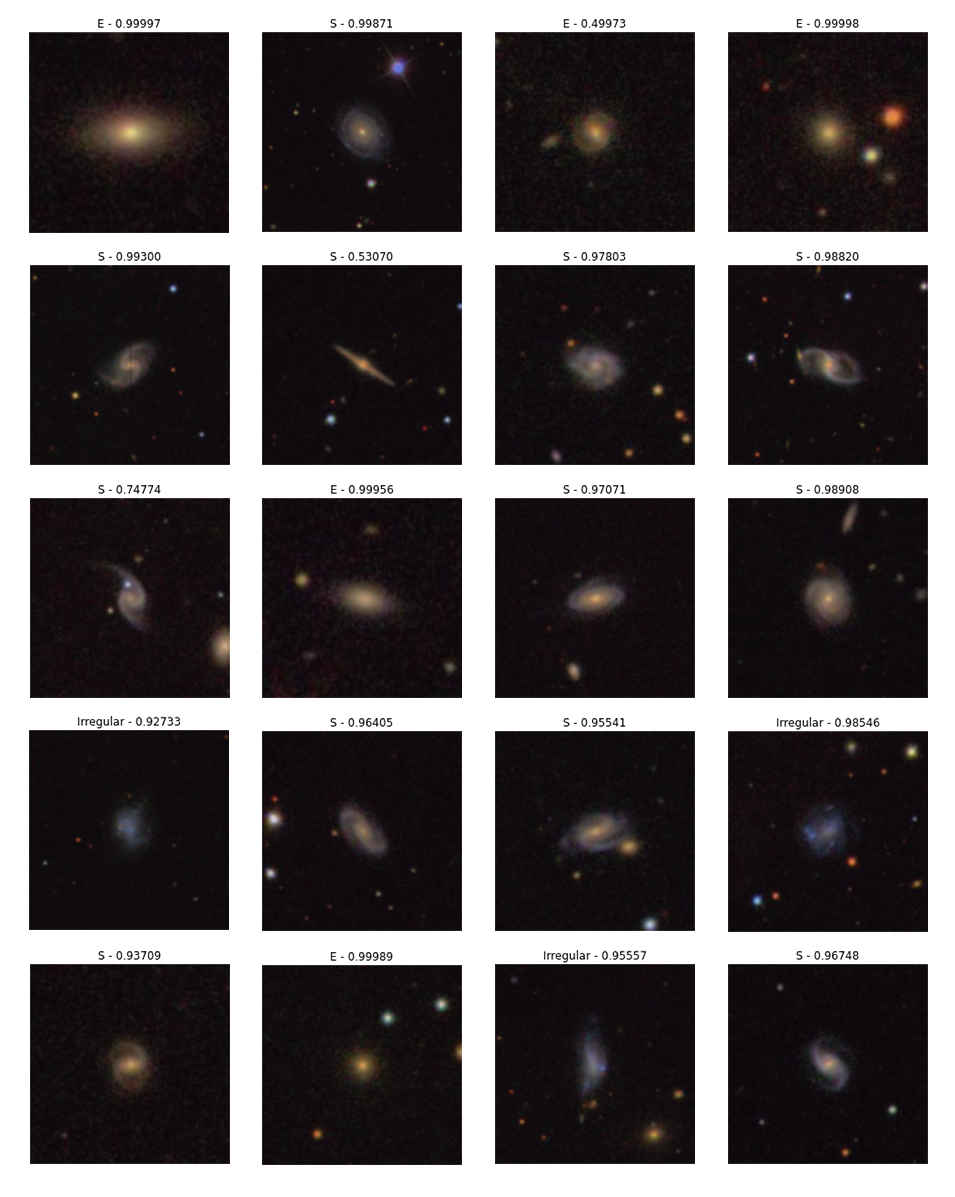} 
\caption{Examples of images from Kaggle dataset as predicted by SpinalNet with their labels and their probabilities.}
\label{fig2}
\end{figure}

At our first runs we have obtained an accuracy of around 97\%, but we found that, the result is not conclusive as most of image in the dataset were taken are elliptical galaxies which were predicted more effectively, comparing them with the other classes of the galaxies in our datasets. Then we tried to align them to have comparably the same size. A total of 4,564 images were used for classification. Fig. \ref{fig2} shows some typical examples of images taken from Kaggle datasets as predicted by SpinalNet with their corresponding class labels and their probabilities acquired from the voting.

But we have to state that we had data limitations of our image dataset and therefore some poorly represented types of galaxy types, such as irregular ones, with even setting lowest possible threshold hardly could have same amount of class images.

Irregular galaxies are also hardly recognized by non-expert volunteers involved in Galaxy Zoo project. Same problem was raised in assessing data quality in Citizen Science, where data collected using crowd-sourced scientific research method instead of hiring a group of experts. Not all crowd-sourced research projects can involve the help of non-expert volunteers. One example can be a situation when they have to apply series of methods or perform repetitive tasks for a longer period of time. In those kinds of cases, untrained or non-expert human resources used in project may have a high risk of corrupting the data \cite{Thelen2008}. Thus, we have manually removed the false positive images in order to achieve sensible results.


\section{Implementation of the SpinalNet on the Galaxy Zoo Datasets}

Recent researches show that image classification is done by NN architectures have high accuracy even on small imagesets. Some of these works include \cite{Kabir, Kim, KHe2016, Perez2017, Peth2016, Cheng2020, Dominguez2019}. SpinalNet algorithm for example is an architecture which mimics the natural way of reacting \cite{Kabir}. We choose it among all others because it has been implemented on various benchmark datasets and proved to give the state-of-the-art performance, more over we found their code to be simple to use. Among all implementations we have tested out we picked up a simplest and the fastest one which uses PyTorch library. We observed that by comparing the implementations using Tensorflow with that of PyTorch, PyTorch can work with new CUDA API and exhibit backward compatibility.

\begin{figure}
\centering
\includegraphics[width=0.43\textwidth]{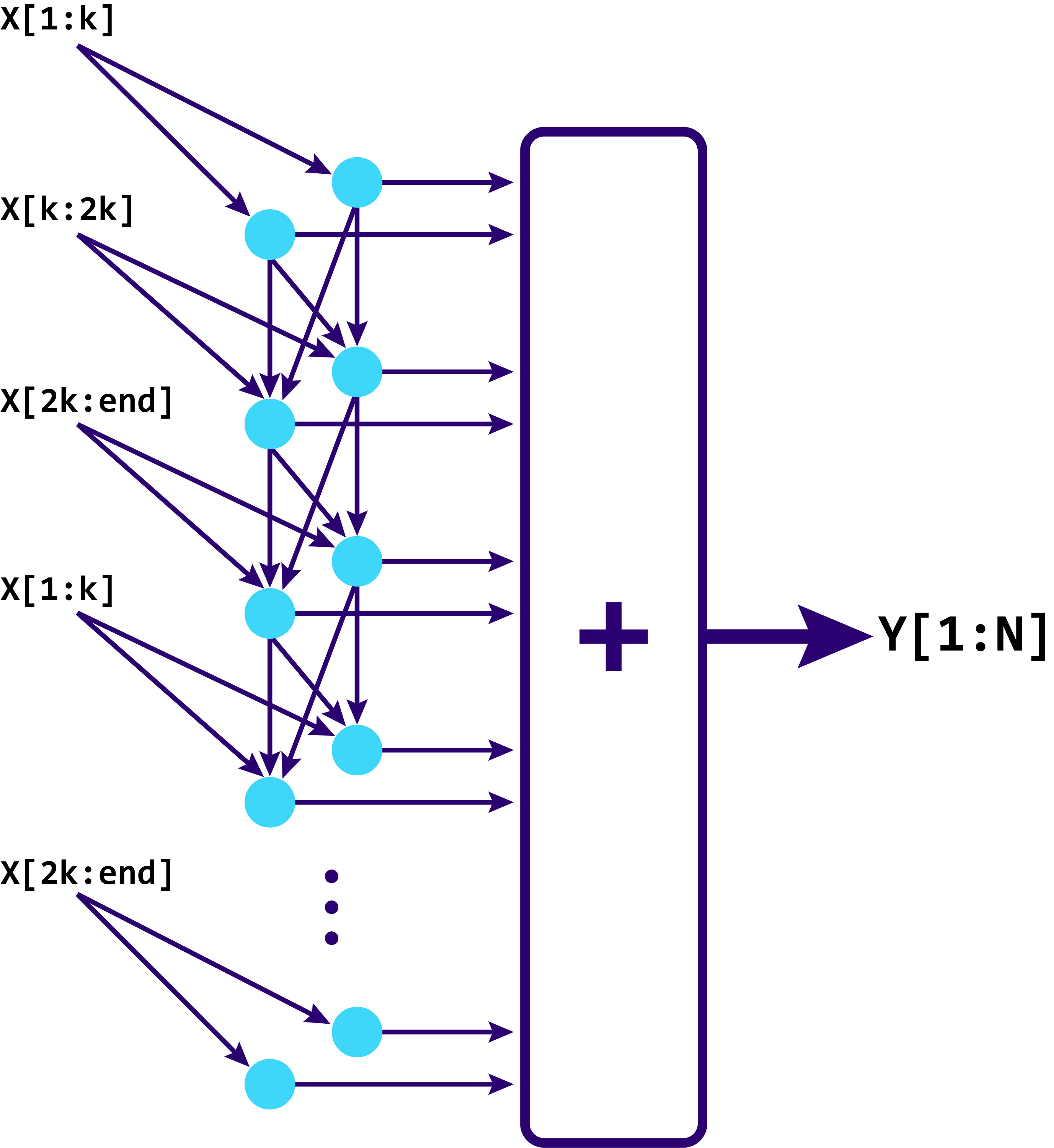} 
\caption{The architecture of the used DNN.}
\label{fig3}
\end{figure}

Assuming all above SpinalNet was used to give another tryout to solve galaxy classification problem. SpinalNet is a deep neural network, the architecture of which is shown in Fig. \ref{fig3}. The proposed by \cite{Kabir} Neural Network consists of the input row, the intermediate rows of hidden layers, and the output row.

Furthermore, step by step training may allow us to gradually expand SpinalNet depth. Number of addition neurons are specified by number of classes passed as an input. In the middle row we have several hidden layers. As it is known, hidden layers accept a portion of input, and all layers not including the first one have outputs as an input data from prior layers. As a result, the output layer tots up all hidden neurons' outputs of the medium row. Depending on the number of classes, in our case we have 2, 3 and 10, architecture supposes to have exact same number of input nodes. Number of features will affect on how many hidden layers a given DNN will have. Eventually, the output layer has the same number of outputs as the number of provided galaxy data classes. These classes are: Elliptical and Spherical; E, S and Irr; E0, E3, E7, SBa, SBb, SBc, Sa, Sb, Sc, Irr. Nice side of this architecture is the fact that it discards irrelevant data. In some cases, data reduction can lead to a decrease in efficiency. Increasing the processing power will not lead to significant efficiency improvement. For all our experiments and tests we have used PyTorch version of SpinalNet code with version 1.9.1 of torch library and to decrease processing time we used CUDA Toolkit version 11.1.1. PC specifications used for code running were - CPU AMD Ryzen 7 5800H 3.20GHz 8 cores, GPU NVIDIA GeForce RTX 3060, RAM 16GB.

\section{Results and Discussions}

We have divided the data into the training and testing sections. Further training and testing data were divided into sub-folders according to the number of classes we were going to recognize. As a part of our results, we would like to show and discuss Fig. \ref{fig2}. We can see that overall accuracy is very high and, in most cases reach values of 98\% even having extraneous artifacts around target galaxy (i.e., row 1, col 4; row 2, col 3; row 5, col 2). However, you can mention examples what have been predicted incorrectly - spiral galaxy in first row was predicted as elliptical with predicted value 0.49973. Another interesting case we found is a side-turned spiral galaxy (row 2, col 2) which have been predicted with a low value of 0.53070 - this example may mean that SpinalNet needs additional features to be added, or that imageset needs more examples describing this case.

With a high degree of confidence, we can state that this DNN is the most effective tool to classify galaxies according to their morphology. Starting with the Kaggle classes E and S, we can confidently state that we can recreate the human eye classification with all methods and samples (accuracy $>$ 98\%) as shown in Table \ref{table:1}. The task becomes much more complex when trying to discriminate between 10 classes, as it would be not that easy for a human eye, and the best result is 82\% using SpinalNet. However, if we simply utilize three classifications, elliptical (E), spiral (S) and irregular (I) galaxies, we find an accuracy $>$ 95\% with the DNN for all data.

We have achieved remarkable results classifying on 2, 3 and 10-class datasets using SpinalNet architecture. Outperforming best result configuring hyperparameters of the Neural Network we could reach best values on all classes comparing to results reported in \cite{Cavanagh2021} which shown an  accuracies of 92\%, 82\% and 77\% for the binary, 3-way and 4-way classifications respectively.




\begin{table}[t]
\caption{Accuracy for the used deep neural network for 10 epochs.}
\begin{center}
 \begin{tabular}{|c | c | c | c | c |} 
\hline
\textbf{ N} & \textbf{8} & \textbf{16} & \textbf{24} & \textbf{32} \\ [0.5ex] 
 \hline
 \textbf{2 classes} &  0.962934 & 0.970820 & 0.966088 & 0.981861 \\ 
  \hline
 \textbf{3 classes} & 0.923741 & 0.948921 & 0.946763 & 0.943165 \\ 
  \hline
 \textbf{10 classes} & 0.689209 & 0.782014 & 0.790647 & 0.818705 \\ 
 \hline

\end{tabular}
\label{table:1}
\end{center}
\end{table}

\begin{figure}[b]
\centering
\includegraphics[width=0.45\textwidth]{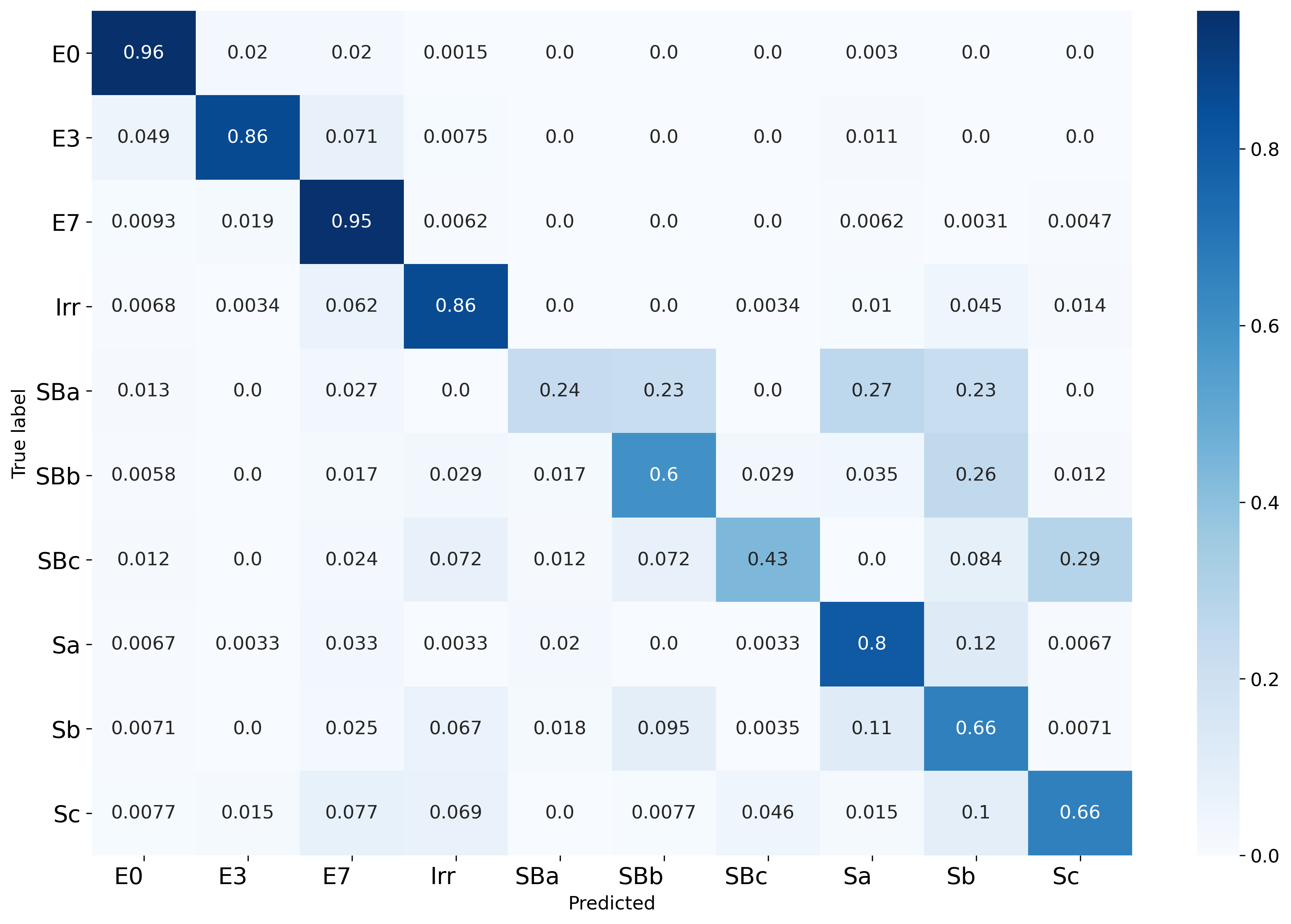} 
\caption{Confusion matrix for the 10-class classification.}
\label{fig4}
\end{figure}

Fig. \ref{fig4} represents confusion matrix for 10-class classification case which shows the percentage of correctly predicted objects in each class. In the best-case scenario, the confusion matrix will be filled with ones only along the diagonal, and with zeros otherwise. But as we can see SpinalNet being highly confused predicting barred spiral galaxies. In some "barred" cases false-negative results. For instance, we can see that SBb has been classified wrongly as Sb. This gives us a room for improvement for future research work.

\section{Conclusion}

In this work we have implemented a galaxy classification algorithm based on a deep neural network (DNN) with a gradual input. Original architecture known as SpinalNet \cite{Kabir}, which mimics somatosensory system of human, was used on a preprocessed Galaxy Zoo data. Having performed a huge number of tasks we are able to prepare the image datasets which would perfectly fit the needs of the project. Eventually, we could reach a significant error reduction running the image classification process on a single laptop with CUDA framework, which proves that our approach could be reproduced even on a domestic computational unit. Applying it to the Galaxy Zoo dataset we are able to distinguish between the two main classes of galaxies namely, elliptical, spiral above 98\% and (+irregulars) close to 95\% and 10-class 82\%.


\section*{Acknowledgment}

D. Shaiakhmetov and F. Unsal would like to give thanks to the Ala-Too International University for the scholarship opportunities which is provided to them. We would like to thank Dr. S. Cankurt and Dr. D. Kabir for the fruitful discussions we had on this work.


\begin{thebibliography}{00}
\bibitem{Kabir} Kabir H. M., Abdar M., Jalali S. M. J., Khosravi A., Atiya A. F., Nahavandi S., \& Srinivasan D. (2020). Spinalnet: Deep neural network with gradual input. arXiv preprint arXiv:2007.03347., unpublished.


\bibitem{Martin2020}G Martin, S Kaviraj, A Hocking, S C Read, J E Geach, Galaxy morphological classification in deep-wide surveys via unsupervised machine learning, Monthly Notices of the Royal Astronomical Society, Volume 491, Issue 1, January 2020, Pages 1408–1426, https://doi.org/10.1093/mnras/stz3006

\bibitem{GalaxyZooWebPage} Kaggle ``Galaxy zoo challenge" URL:https://www.kaggle.com/c/galaxy-zoo-the-galaxy-challenge/overview/about-galaxy-zoo (visited on 15/12/2021)


\bibitem{Willett2013} Kyle W. Willett, Chris J. Lintott, Steven P. Bamford, Karen L. Masters, Brooke D. Simmons, Kevin R. V., et al., “Galaxy Zoo 2: detailed morphological classifications for 304 122 galaxies from the Sloan Digital Sky Survey”, Monthly Notices of the Royal Astronomical Society, vol. 435, no. 4, pp. 2835–2860, Nov. 2013, doi: 10.1093/mnras/stt1458.


\bibitem{Calleja2004} Jorge De La Calleja, Olac Fuentes, Machine learning and image analysis for morphological galaxy classification, Monthly Notices of the Royal Astronomical Society, Volume 349, Issue 1, March 2004, Pages 87–93, https://doi.org/10.1111/j.1365-2966.2004.07442.x


\bibitem{Kalvankar2020} Kalvankar, Shreyas, Hrushikesh Pandit, and Pranav Parwate. "Galaxy morphology classification using efficientnet architectures." arXiv preprint arXiv:2008.13611 (2020), unpublished.


\bibitem{Cavanagh2021} Cavanagh, Mitchell K., Kenji Bekki, and Brent A. Groves. "Morphological classification of galaxies with deep learning: comparing 3-way and 4-way CNNs." Monthly Notices of the Royal Astronomical Society (2021), in press.


\bibitem{Manda2010} Manda Banerji, Ofer Lahav, Chris J. Lintott, Filipe B. Abdalla, Kevin Schawinski, Steven P. Bamford, et al., Galaxy Zoo: reproducing galaxy morphologies via machine learning, Monthly Notices of the Royal Astronomical Society, Volume 406, Issue 1, July 2010, Pages 342–353, https://doi.org/10.1111/j.1365-2966.2010.16713.x

\bibitem{Kuminski2014} Evan Kuminski, Joe George, John Wallin, Lior Shamir, "Combining human and machine learning for morphological analysis of galaxy images." Publications of the Astronomical Society of the Pacific 126.944 (2014): 959.

\bibitem{Variawa2020} M. Z. Variawa, T. L. van Zyl and M. Woolway, "A rules-based and Transfer Learning approach for deriving the Hubble type of a galaxy from the Galaxy Zoo data," 2020 IEEE 23rd International Conference on Information Fusion (FUSION), 2020, pp. 1-7, doi: 10.23919/FUSION45008.2020.9190462.

\bibitem{Mike2021} Mike Walmsley, Chris Lintott, Tobias Géron, Sandor Kruk, Coleman Krawczyk, Kyle W Willett, et al., Galaxy Zoo DECaLS: Detailed visual morphology measurements from volunteers and Deep Learning for 314,000 galaxies, Monthly Notices of the Royal Astronomical Society, 2021;, stab2093, https://doi.org/10.1093/mnras/stab2093

\bibitem{Hubble} Hubble, Edwin Powell. The realm of the nebulae. Vol. 25. Yale University Press, 1982.

\bibitem{Thelen2008} Thelen, Brett \& Thiet, Rachel. (2008). Cultivating connection: Incorporating meaningful citizen science into Cape Cod National Seashore's estuarine research and monitoring programs. Park Science. 25. 

\bibitem{Kim} Kim, E. J. \& Brunner, R. J. Star–galaxy classification using deep convolutional neural networks. Mon Not R Astron Soc 464, 4463–4475 (2017).

\bibitem{KHe2016} K. He, X. Zhang, S. Ren and J. Sun, "Deep Residual Learning for Image Recognition," 2016 IEEE Conference on Computer Vision and Pattern Recognition (CVPR), 2016, pp. 770-778, doi: 10.1109/CVPR.2016.90.

\bibitem{Perez2017} Wang, Jason, and Luis Perez. “The Effectiveness of Data Augmentation in Image Classification Using Deep Learning.” Convolutional Neural Networks Vis. Recognit 11 (2017): 1-8.

\bibitem{Peth2016} Peth, Michael. (2016). Using Machine Learning to Study the Relationship Between Galaxy Morphology and Evolution. 10.5281/zenodo.57549. 

\bibitem{Cheng2020} Ting-Yun Cheng, Christopher J Conselice, Alfonso Aragón-Salamanca, Nan Li, Asa F L Bluck, Will G Hartley, et al., “Optimizing automatic morphological classification of galaxies with machine learning and deep learning using Dark Energy Survey imaging,” Monthly Notices of the Royal Astronomical Society, vol. 493, no. 3, pp. 4209–4228, Apr. 2020, doi: 10.1093/mnras/staa501.

\bibitem{Dominguez2019} H Domínguez Sánchez, M Huertas-Company, M Bernardi, D Tuccillo, J L Fischer, Improving galaxy morphologies for SDSS with Deep Learning, Monthly Notices of the Royal Astronomical Society, Volume 476, Issue 3, May 2018, Pages 3661–3676, https://doi.org/10.1093/mnras/sty338

\end{thebibliography}
\end{document}